\documentclass[12pt]{elsarticle}

\usepackage{amssymb}
\usepackage{xcolor}
\usepackage{amsmath}
\usepackage{hyperref}
\usepackage{geometry}
\usepackage{booktabs}
\usepackage{longtable}
\usepackage{pdflscape}
\usepackage{geometry}
\usepackage{caption}
\usepackage{lineno}
\usepackage{eso-pic}

\AddToShipoutPictureFG*{%
  \AtPageUpperLeft{%
    \put(0,-50){%
      \parbox{\paperwidth}{%
        \centering
        \small\textit{ This is a preprint for the manuscript published in the Journal of Imaging Informatics in Medicine.\\ 
        The final version of record may differ slightly from this version and is available at: \url{https://link.springer.com/article/10.1007/s10278-026-01958-4}\\ A formatted preview by Springer Nature is available at: \url{https://rdcu.be/fet5p}}
      }%
    }%
  }%
}
\begin{document}

\begin{frontmatter}



\title{Rethinking Pulmonary Embolism Segmentation: A Study of Current
Approaches and Challenges with an Open Weight Model}


\author[label1]{Yixin Zhang}
\author[label2]{Ryan Chamberlain}
\author[label3]{Lawrence Ngo}
\author[label2]{Kevin Kramer}
\author[label1]{Maciej A. Mazurowski}

% Author affiliation
\affiliation[label1]{organization={Duke University},%Department and Organization
            city={Durham},
            state={NC},
            country={USA}}

\affiliation[label2]{organization={Minnesota Health Solutions},%Department and Organization
            % addressline={Minneapolis}, 
            city={Minneapolis},
            state={MN},
            country={USA}}
            
\affiliation[label3]{organization={CoRead.ai},%Department and Organization
            city={Durham},
            state={NC},
            country={USA}}

%% Abstract
\begin{abstract}
\textcolor{black}{
Pulmonary Embolism (PE) is a life-threatening condition for which accurate and timely detection is critical to patient care. However, our systematic study of PE segmentation algorithms reveals concerning limitations in the current state of research. Challenges such as small and inconsistent datasets, a lack of reproducible baselines, and limited comparative evaluation across models are hindering progress in the field.
In this study, we curated a densely annotated dataset comprising 490 CTPA scans, each from a unique patient (430 for training and 60 for testing). We evaluated nine widely used segmentation architectures, including both CNN- and ViT-based models, in 2D and 3D configurations, using mean Dice Similarity Coefficient (mDSC) and Average Symmetric Surface Distance (ASSD) as evaluation metrics. Furthermore, the highest-performing model was evaluated on a public dataset without fine-tuning and achieved reasonable generalization performance.
Our results show that: (1) a 3D U-Net with ResNet encoding blocks remains a highly effective architecture for PE segmentation; (2) 3D models consistently outperform their 2D counterparts; (3) across all architectures, when trained and evaluated on the same datasets, model error patterns are highly consistent; and (4) distal emboli remain particularly challenging due to both task complexity and the scarcity of high-quality datasets, highlighting the need for datasets with more comprehensive and consistent distal PE coverage.
To promote research reproducibility, the architecture and pretrained weights of our best-performing model are publicly available at: }
\url{https://github.com/mazurowski-lab/PulmonaryEmbolismSegmentation}

\end{abstract}


% \begin{highlights}
% \end{highlights}

%% Keywords

\end{frontmatter}



\section{Introduction}
\label{sec:intro}
Pulmonary embolism (PE) is \textcolor{black}{a major cardiovascular condition associated} with substantial morbidity and mortality, \textcolor{black}{considered a leading cause of cardiovascular death, ranking behind myocardial infarction and stroke.} \cite{afifi2026national} 
\textcolor{black}{In 2019, there were an estimated 393000 cases of PE in the United States.} \cite{doi:10.1161/CIR.0000000000001123} 
Acute PE, when causing hemodynamic instability, is immediately life-threatening and needs to be treated with anti-coagulation and sometimes thrombolytic therapy in a timely manner. \cite{konstantinides20202019} 
Chronic PE, on the other hand, is a causative precursor of thromboembolic pulmonary hypertension (CTEPH) \cite{kim2019chronic}\cite{humbert20222022}. 
Computed tomography pulmonary angiography (CTPA) is the clinical gold standard for diagnosing pulmonary embolism. 
The development of algorithms and models for identifying and characterizing both acute and chronic emboli from CTPA is hence of high clinical relevance. Voxel-level segmentation of PE from CTPA provides detailed information by characterizing the location, morphology, and \textcolor{black}{size} of emboli.
This enables the quantification of clot burden, which leads to a more precise assessment of disease severity for better-informed treatment decisions.

\textcolor{black}{There is a large body of prior studies on PE segmentation. 
Arabian et al. \cite{11213757} achieved a patient-level Dice score of 0.5 with their proposed model when trained and evaluated on the FUMPE \cite{masoudi2018new} and CAD-PE \cite{CAD-PE} datasets. 
Cano-Espinosa et al. \cite{cano2020computer} proposed a multi-slice, multi-axial model, achieving a per-embolus sensitivity of 0.68 at approximately one false positive per scan, evaluated on 20 cases from a public dataset. 
Cheng et al. \cite{cheng2023feature} introduced a feature-enhanced, adversarial semi-supervised network based on HRNet, demonstrating performance gains with unlabeled data. 
Djahnine et al. \cite{djahnine2024detection} presented a 3D pipeline that jointly performed clot segmentation and Qanadli score estimation, reporting $R^2 = 0.72$ for severity prediction. 
Do˘gan et al. \cite{dougan2024enhanced} trained their variant of Mask R-CNN on 36 patients and reported a DSC of 0.95 on a separate test set of 14 patients from their in-house dataset. However, they did not report the composition of their data. Given the visualization, their data likely contained only central and lobar PE, which may explain their near-perfect results. 
Liu et al. \cite{liu2022cam} proposed CAM-WNet with coordinate attention and pyramid pooling, while Tang et al. \cite{tang2022pulmonary} incorporated CBAM attention; both reported high Dice scores on small in-house datasets ($n \leq 25$). 
Liu et al. \cite{liu2020evaluation} reported high AUC scores on several hundred cases but did not include Dice metrics, limiting assessment of segmentation quality.
Munir et al. \cite{Munir2025DAUNetAL} proposed DAUNet by incorporate deformable convolution mechanism and simAM, an form of attention block on skip connection. Their model achieved a DSC of 0.888 on FUMPE data.   
Kahraman et al. \cite{kahraman2024enhanced} selected and annotated 149 PE-positive CTPAs and trained models in conjunction with 551 PE-negative CTPAs. 
Zhan et al. \cite{Zhan2024BFNetAF} proposed a modification to the U-Net architecture by adding a multi-hierarchical feature fusion layer.}

\textcolor{black}{Besides fully supervised training with dense annotations, another line of work in PE segmentation explores weakly supervised learning strategies. 
Condrea et al. \cite{condrea2024label} applied explainability maps to iteratively refine labels derived from classification outputs. 
Pu et al. \cite{pu2023automated} trained a CNN on high-confidence embolus candidates derived from vascular segmentation. 
Yang et al. \cite{yanggraph} used graph-cut pre-segmentation to generate pseudo-labels. Although these weakly supervised methods are innovative, they may introduce biases when encountering difficult cases such as distal emboli or anatomically ambiguous regions, which could limit their long-term clinical utility.}

\textcolor{black}{However, despite the large body of prior work, the current research landscape in PE segmentation remains fragmented and faces significant challenges.}
The first challenge arises from the limitations of public datasets. As of 2025, three datasets and their derivatives for PE segmentation are publicly available: FUMPE (35 scans) \cite{masoudi2018new}, CAD-PE (91 scans) \cite{CAD-PE}, and READ (40 scans) \cite{de2023pixel}. 
The three datasets are annotated with different criteria: FUMPE considers only central, lobar, and segmental PE as annotation targets, omitting the sub-segmental ones. CAD-PE shows a high tendency to mark partial-volume artifacts in thin arteries and veins as distal PE. Furthermore, it occasionally annotates the vessel with emboli rather than the emboli themselves. 
READ is an accurately annotated dataset for acute pulmonary embolism (PE) segmentation at all anatomical levels. It contains 20 CTPAs acquired on Toshiba and 20 on GE CT scanners with near-isotropic, sub-millimeter resolution. This high imaging resolution allows small distal PE, which are otherwise invisible on routine CTPA, to be annotated.    
Unfortunately, each dataset alone is relatively small, making it challenging to withhold a sufficiently large subset for testing after the train-val-test split. On the other hand, training models on one dataset while testing on another is impractical due to the inconsistent PE annotation criteria.         

The second challenge arises from the non-standard metric reporting and data handling protocols in much of the prior literature. \textcolor{black}{Regarding metric reporting, some} studies report slice-level ROC--AUC rather than DSC as the primary evaluation metric for their segmentation models \cite{liu2020evaluation}\cite{condrea2024label}. This practice does not reflect the volumetric nature of 3D segmentation tasks. It also violates the independent sampling assumption \textcolor{black}{that is} vital to the ROC-AUC metric and is prone to result distortion due to class imbalance \cite{metric_rigorness}.
\textcolor{black}{Some other studies used pixel-wise annotations for patient-level classification, rather than semantic segmentation \cite{kahraman2024enhanced}, and consequently did not report segmentation metrics such as DSC and ASSD.}
\textcolor{black}{Regarding data handling, it is concerning to observe data leakage appearing frequently in works proposing novel model architectures \cite{liu2022cam}\cite{tang2022pulmonary}\cite{Munir2025DAUNetAL}\cite{trongmetheerat2023segment}. In those works,} slices in the test set are selected from the same scans as those appearing in the training set, leading to inflated model performance. \textcolor{black}{These flaws are particularly likely to go unnoticed, especially when only a few studies have made their code accessible for reproducibility and none have published their model weights for out-of-box usage.}

To alleviate these gaps and clarify the research landscape of PE segmentation, we present an empirical study that leads to:
\begin{enumerate}
    \item \textbf{An Open-Sourced Model with Weights:} We develop and release the first open-weight PE segmentation models trained on a large, high-quality in-house dataset. The dataset contains 490 manually annotated CTPA scans (430 for training, 60 for testing) with acute and chronic PE annotated across all anatomical levels. Each scan is obtained from a unique patient diagnosed as PE positive. Each patient is randomly sampled from the database of collaborating institutions.
    
   \item \textbf{A Benchmarking Study for Segmentation:} We implement, train, and evaluate nine representative segmentation architectures under a consistent pipeline. We then report the \textcolor{black}{average Symmetric Surface Distance (ASSD)}, the mean Dice Similarity Coefficient (DSC), and the per-patient DSC for patients in the test set \textcolor{black}{for each trained model}. Our results show that patients in the test set are ranked consistently by their DSC scores across different models, regardless of architecture. This observation suggests an intrinsic difficulty stratification within the dataset. We further analyze the factors underlying this stratification.
    
    \item \textbf{A Generalization Study through Detection:} 
    Ideally, external PE segmentation datasets should be used to evaluate model generalizability. However, differences in annotation criteria and dataset limitations can introduce more noise than useful information if used directly. Since a PE segmentation model can be naturally repurposed for PE detection, we instead perform a detection-based evaluation as a proxy for generalization, whose statistics also allow us to examine potential domain shifts when data are curated from different institutions.

\end{enumerate}
\section{Materials and Methods}
\subsection{Study Design}
\color{black}
In addition to the literature review, our study contains three experiments to evaluate model architecture, pretraining, and generalization for PE segmentation on CTPA images:
\begin{enumerate}
    \item \textit{Architecture Benchmarking:} The 490 annotated in-house CTPA volumes are split into training and test sets. Nine segmentation architectures are trained under a unified protocol to assess architectural suitability and characterize error patterns across models.
    \item \textit{Effect of Pretraining:} Based on observations from \textit{Architecture Benchmarking}, we select Inception U-Net (the best-performing 2D model). The model is pretrained on the RSNA-PE dataset ($>$7{,}000 volumes with slice-level PE labels) and fine-tuned on varying subsets of the in-house data to evaluate performance gains and their dependence on training set size.
    \item \textit{Model Generalization:} The best-performing model from \textit{Architecture Benchmarking} is evaluated on public data without fine-tuning to assess out-of-the-box performance. Performance differences relative to the in-house test set are analyzed to identify potential sources.
\end{enumerate}

These experiments, in conjunction with our preliminary audit of the literature on metric reporting and evaluation practices, provide a systematic assessment of the current landscape of PE segmentation studies.

\color{black}
\subsection{Dataset}
\subsubsection{In-House Dataset for PE Segmentation}
Our study utilized 490 de-identified CTPAs (one per patient) with confirmed PE. Voxel-level embolus annotations were derived from radiology reports generated by board-certified radiologists.
The scans were collected from multiple institutions and encompassed a variety of scanner models, contrast agents, and clinical protocols. \textcolor{black}{More details about the imaging parameters can be found in \ref{Image_param}.} Patient age was available in the de-identified metadata (mean $61.3 \pm 16.1$ years). 
All CTPAs were converted from DICOM to NIfTI format with RAS orientation, and a research assistant visually verified the spatial consistency between images and annotations. 
The dataset was split into 430 scans for training/validation and 60 scans for a held-out test. \textcolor{black}{(i.e., 7.17:1 trainval-test split)}. In the test sets, an average of 3.483 thrombus fragments per CTPA is inferred from annotation. 

\subsubsection{Public Datasets for Detection Validation}
\label{sec:public_validation}
\begin{figure}[h!]
    \centering
    \includegraphics[width=.75\linewidth]{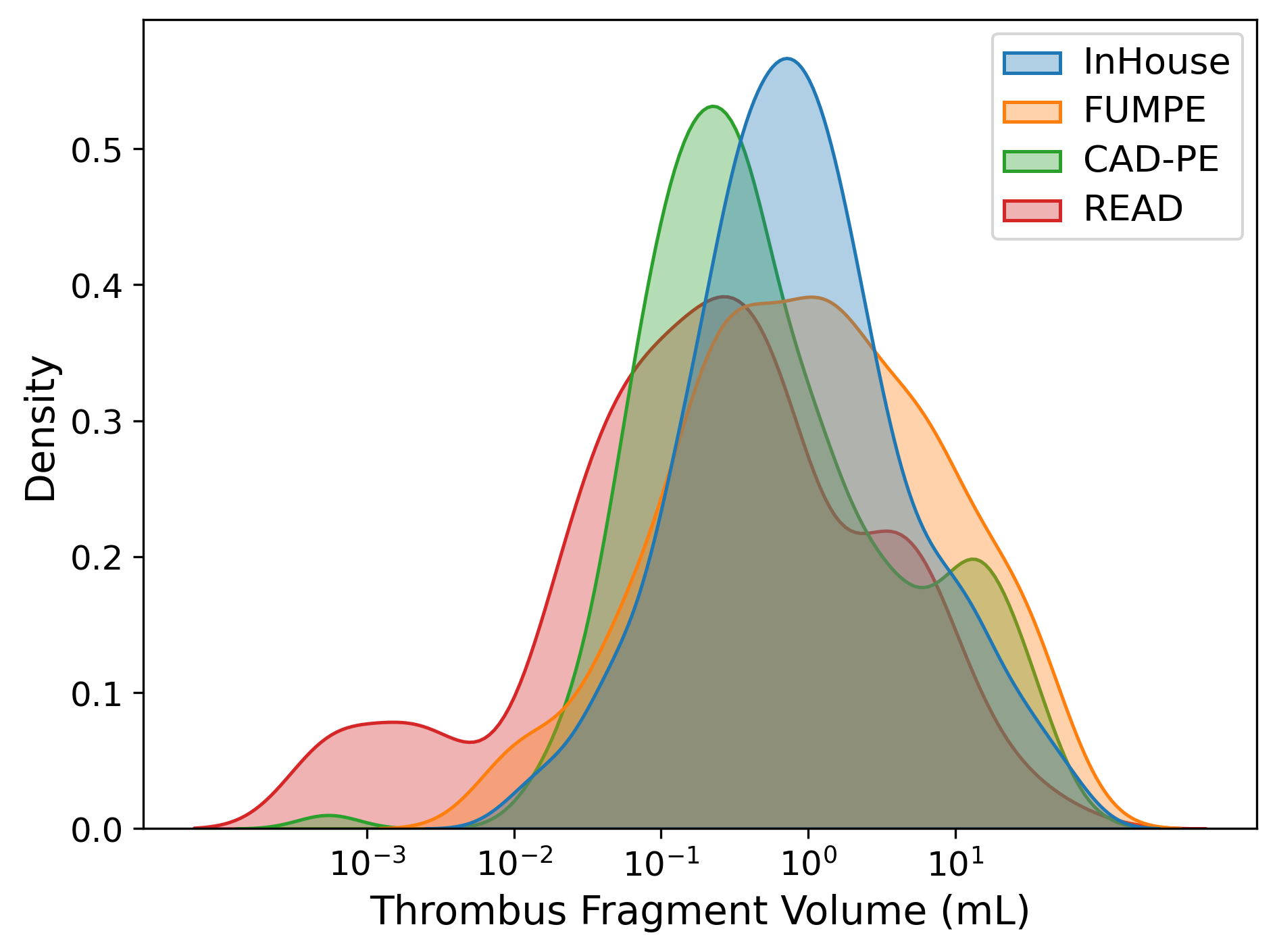}
    \caption{Across all four datasets, the volume distribution of thrombus fragments is consistent with a log-normal distribution.}
    \label{fig:thrombus_size_dist}
\end{figure}

To validate whether our model generalizes to external datasets, we observe its performance on three public PE datasets: FUMPE \cite{masoudi2018new}, CAD-PE \cite{CAD-PE}, and READ \cite{de2023pixel}. For these datasets, an average of \textcolor{black}{2.66} thrombus fragments per CTPA is annotated for FUMPE, \textcolor{black}{4.83} for READ, and \textcolor{black}{4.21} for CAD-PE. Figure \ref{fig:thrombus_size_dist} shows the distribution of the volume of \textcolor{black}{thrombus fragments} in the three public datasets, along with that of our in-house dataset.

The distribution patterns align with the statement in Section \ref{sec:intro} that these datasets differ substantially in annotation standards, which challenges the effectiveness of the mDSC metric. To circumvent this issue, we instead measure the performance of thrombus fragment detection. By analyzing the counts and sizes of true positives (TP), false positives (FP), and false negatives (FN) fragments, we can verify whether the model's generalizability is justifiable. We present and interpret the detection \textcolor{black}{metrics} in Section \ref{metric:detect} and the results in Section \ref{sec: rslt_generalizability}.
%-------------------------------------------------------------------------
\subsection{Evaluation Metrics}
When computing the evaluation metrics, the native annotations provided by the dataset are always used as the ground truth without applying any spatial transformations.

\subsubsection{Mean Dice Similarity Coefficient (mDSC)}
The segmentation performance of all trained models was evaluated per CTPA using the Dice Similarity Coefficient \cite{sorensen1948method}:
\begin{equation}
  DSC = \dfrac{2TP}{2TP+FP+FN}
  \label{eq:dsc}
\end{equation}
where $TP$, $FP$, and $FN$ are voxel counts. The final metric was the mean DSC across all test CTPAs.

\subsubsection{Average Symmetric Surface Distance (ASSD)}
\textcolor{black}{ASSD \cite{heimann2009statistical} quantifies boundary accuracy and provides complementary insights beyond overlap-based metrics such as DSC, which may fail to capture boundary discrepancies. Mathematically, ASSD takes the ground-truth mask and the predicted segmentation mask as input and is defined as:
\begin{equation}
ASSD(S_P, S_G) =
\frac{1}{|S_P| + |S_G|}
\left(
\sum_{x \in S_P} \min_{y \in S_G} \|x - y\|
+
\sum_{y \in S_G} \min_{x \in S_P} \|y - x\|
\right)    
\label{eq:assd}
\end{equation}
Although ASSD does not explicitly account for false positives (FP) and false negatives (FN) at the object level, and may become less informative under substantial variability in annotation standards, we include it for completeness. In reporting ASSD, we exclude FP and FN regions from the metric computation and instead report their counts separately to remain conceptually aligned with the mathematical definition of the metric.}
\subsubsection{Embolus-level detection analysis}
\label{metric:detect}
In addition to computing voxel-level DSC, we also performed embolus-level detection analysis to enable the evaluation described in Section \ref{sec:public_validation}.
Many prior studies define successful detection as "having one-pixel overlap between prediction and ground truth". However, this definition may exaggerate detection performance by accepting accidental overlaps. 
Instead, we adopt a stricter, volume-aware criterion. Specifically, we first apply morphological operations to identify individual emboli in both the human annotations and model predictions. Then, two stages of screening under a detection threshold $X\%$ are applied to characterize the FPs and FNs. 

In the first stage, each predicted embolus is checked for their overlaps with emboli from the annotation. If less than $X\%$ of its volume overlaps with true emboli, the prediction is marked as a false positive (FP); otherwise, it is a true positive (TP). In the second stage,  each annotated embolus is verified. If less than $X\%$ of its volume is covered by TPs, meaning its mass is not sufficiently flagged by the model, it will be considered a missed embolus (FN). 
This approach deprecates both incidental touch and excessive over-segmentation, providing a more robust evaluation. 

\subsection{Data Preprocessing and Augmentation}
For 2D models, each CTPA volume is sliced along the axial plane without spatial resampling. This gives an average spacing of (0.752 $\pm$ 0.097, 0.752 $\pm$ 0.097, 2.216 $\pm$ 1.011) mm/voxel. For 3D models, all CTPAs were resampled to a consistent voxel spacing of (0.7373, 0.7373, 1.0) mm/voxel. The image intensity was clipped at -195 HU and 310 HU (Hounsfield Units), then centered and normalized following nnU-Net conventions. During training, the data augmentation included random affine scaling (0.8–1.25), random rotations ($\pm20^\circ$ and $90^\circ$), cropping ($384 \times 384$), and horizontal/vertical flips.

\subsection{Model Implementation and Training}
We compared a range of 2D and 3D architectures from both the convolutional neural network (CNN) and vision transformer (ViT) families, with their performance summarized in Table \ref{tab:baseline_comparison}. The model selection was designed to cover both CNN- and ViT-based approaches across 2D and 3D settings, including generic architectures as well as those specifically tailored for PE segmentation.  
For hyper-parameters and architectural details, we followed the configurations provided in the original publications or official repositories whenever available (e.g., Chen et al. \cite{Chen_2024_WACV}). When such references were not available, we re-implemented the models and trained them with reasonable adjustments to ensure stable convergence using AdamW optimizer (learning rate $8\times10^{-4}$, weight decay $1\times10^{-5}$).  A One-Cycle learning rate scheduler is also used, taking 10\% of the training epochs for linear warm-up. A combination of batch-aggregated DiceLoss and CrossEntropyLoss was used as the loss function. 
\textcolor{black}{For SCUNet++, training the model with its original configuration resulted in suboptimal performance on our dataset. 
To ensure a fair comparison, we modified the training pipeline as follows: we extracted 448x448 center crops from the original 512×512 images and resized them to 224×224 before feeding them into the model. Its 224x224 output will then be zoomed to 448x448 for loss computing. 
We optimized the network using AdamW with a peak learning rate of 5e-4 and
weight decay of 0.01, and employed a one-cycle learning rate scheduler with 10\%
of the total training epochs allocated for warm-up.}

\begin{figure}[h]
\centering
\includegraphics[width=\linewidth]{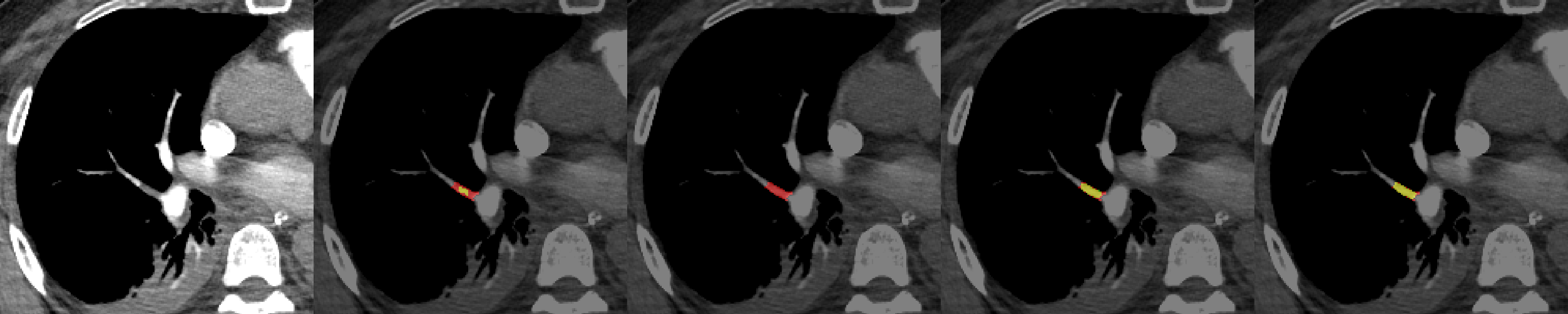}
\includegraphics[width=\linewidth]{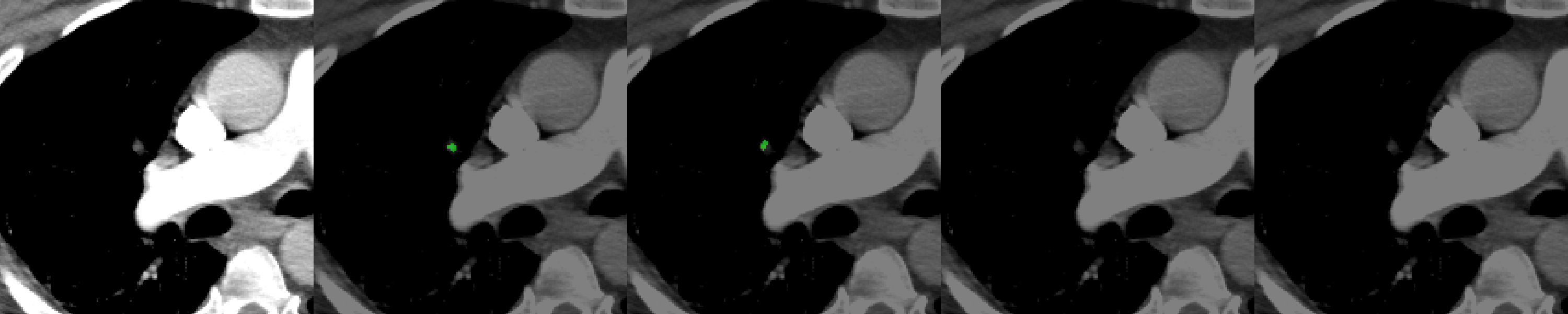}
\includegraphics[width=\linewidth]{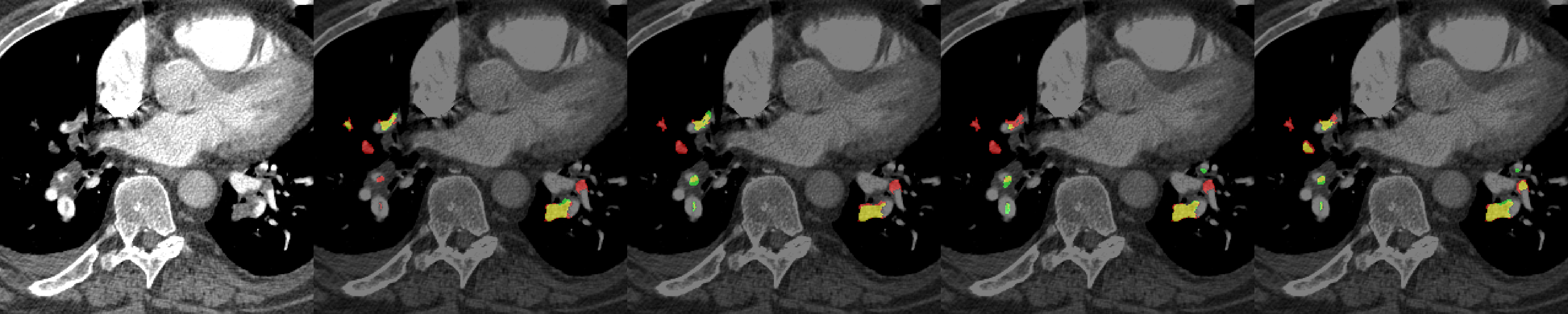}

\begin{tabular}{p{0.175\linewidth} p{0.175\linewidth} p{0.175\linewidth} p{0.175\linewidth}p{0.175\linewidth}}
\centering \textbf{Raw Image} &
\centering \textbf{SegFormer} &
\centering \textbf{Inception} &
\centering \textbf{MedNext3D} &
\centering \textbf{ResUNet3D}
\end{tabular}
\caption{Segmentation results from representative models covering 2D ViT, 2D CNN, 3D ViT and 3D CNN. Red indicates missed ground truth (GT) regions, green denotes over-segmentation, and yellow highlights correctly segmented areas.}
\label{fig:pe-visualization}
\end{figure}

\section{Results}

\subsection{Classification as Pretraining}
\label{sec:pretraining}
\begin{table}[h!]
    \centering
    \begin{tabular}{|c|c|c|}
        \toprule
        \textbf{\# Training Vols.} & \textbf{RSNA Init.}& \textbf{Random Init.}\\
        \midrule
        430 &0.6257$\pm$0.2026 &0.6477$\pm$0.1923 \\
        215 &0.5997$\pm$0.2090 &0.6095$\pm$0.2054\\
        107 &0.5631$\pm$0.2302 &0.5732$\pm$0.2230 \\
        53 &0.4963$\pm$0.2360 & 0.5157$\pm$0.2228\\
        26 &0.4209$\pm$0.2292 &0.4251$\pm$0.2251\\ 
        \bottomrule
    \end{tabular}
    \caption{Dice scores of the 2D Inception U-Net under different initialization schemes, trained with varying numbers of annotated volumes. Each larger training set is a superset of the smaller ones. Pretraining on the RSNA-PE classification task did not yield improvements over random initialization.}
\label{tab:initilization_scaling}
\end{table}
Before venturing into training a model from scratch, we investigate whether existing large-scale PE classification datasets can benefit voxel-level PE segmentation. In this parallel experiment, we adopted a pre-training strategy using the RSNA Pulmonary Embolism dataset (RSNA-PE) \cite{RSNA-PE}, which contains 7,929 CTPA volumes with slice-level PE labels. 
A global max-pooling layer is attached to the logit output to generate slice-level predictions. 
This allows parameters learned for classification to be transferred for segmentation without architectural changes. 
For the same model architecture, we compare the performance of two instances of the model: one initialized with pretrained weights, while the other is randomly initialized. As reported in Table \ref{tab:initilization_scaling}, increasing the number of annotated training volumes consistently improved the mDSC. However, initialization from pretrained weights did not exhibit superior performance than the alternative approach. This reflects the discrepancy in the features that may be learned for PE classification and segmentation. It also justifies our choice to rely primarily on random or ImageNet pretraining (if available) for model initialization in our study.

\subsection{Segmentation Performance for Existing Model Architectures}
\begin{table}[h!]
    \centering
    \begin{tabular}{lcccc}
        \toprule
        \textbf{Model} & \textbf{mDSC $\uparrow$}& \textbf{ASSD $\downarrow$} & \textbf{\#FP } & \textbf{\#FN } \\
        \midrule
SCUNet++ (orig) \cite{Chen_2024_WACV} & 0.411 $\pm$ 0.253 & 4.9 $\pm$ 6.95 & 798 & 22 \\
SCUNet++ (modified) \cite{Chen_2024_WACV} & 0.571 $\pm$ 0.228 & 1.41 $\pm$ 0.83 & 497 & 22 \\
SegFormer-B5 \cite{xie2021segformer} & 0.588 $\pm$ 0.214 & 1.28 $\pm$ 0.59 & 366 & 20 \\
nnUNet2D-ResEncXL \cite{isensee2024nnu} & 0.617 $\pm$ 0.223 & 1.28 $\pm$ 0.69 & 165 & 26 \\
ResAttUNet \cite{chen2018encoder}\cite{oktay2018attention} & 0.636 $\pm$ 0.217 & \textbf{1.05 $\pm$ 0.66} & 531 & 10 \\
CAM-WNET \cite{liu2022cam} & 0.637 $\pm$ 0.200 & 1.11 $\pm$ 0.54 & 207 & 20 \\
\textbf{Inception-UNet \cite{10.1145/3376922}} & \textbf{0.652} $\pm$ \textbf{0.191} & 1.06 $\pm$ 0.50 & 191 & 24 \\
\midrule
3D Swin-UNETR \cite{hatamizadeh2021swin} & 0.625 $\pm$ 0.194 & 1.20 $\pm$ 0.66 & 238 & 32 \\
3D MedNeXt \cite{roy2023mednext} & 0.628 $\pm$ 0.205 & 1.10 $\pm$ 0.64 & 142 & 26 \\
\textbf{nnUNet3D-ResEncXL} \cite{isensee2024nnu} & \textbf{0.713} $\pm$ \textbf{0.165} & \textbf{1.03 $\pm$ 0.67} & 49 & 28 \\
        \bottomrule
    \end{tabular}
    \caption{Performance of all models on the \textbf{in-house test set}. We excluded FP and FN fragments when computing ASSD (in millimeter) and reported the FP and FN count separately.}
\label{tab:baseline_comparison}
\end{table}
We train and test 9 distinct segmentation model architectures, including both 2D and 3D implementations of CNNs and ViTs (Table \ref{tab:baseline_comparison}). In this process, several interesting results emerge. First, at the architectural level, 3D models consistently yield better performance than their 2D counterparts. This may be because 3D representations allow thin emboli positioned perpendicularly to the axial plane to be more visible to the models, which helps reduce false negatives. It is also easier to filter out artifacts from a real embolus when 3D context is available, which helps reduce false positives. On the other hand, within the same spatial dimension, we find that CNN-based models always outperform transformer-based models. Among 2D models, Inception-UNet and CAM-WNet have roughly 0.05 higher mDSC than SCUNet++ and SegFormer-B5. Among 3D models, nnUNet3D achieved substantially higher mDSC than both MedNeXt and Swin-UNETR. Empirical studies attribute this to transformers’ lack of strong inductive biases compared with CNNs, making them less effective at identifying small objects when the training dataset is modest \cite{khan2022transformers}. \textcolor{black}{Qualitative visualizations of segmentation results from representative models (2D ViT, 2D CNN, 3D ViT, and 3D CNN)} are shown in Figure~\ref{fig:pe-visualization}. \textcolor{black}{The images are cropped and magnified for better visualization.}
 
\subsection{Agreement Between Model Predictions}
\label{sec:error_agreement}
\begin{figure}[h!]
    \centering
    \includegraphics[width=\linewidth]{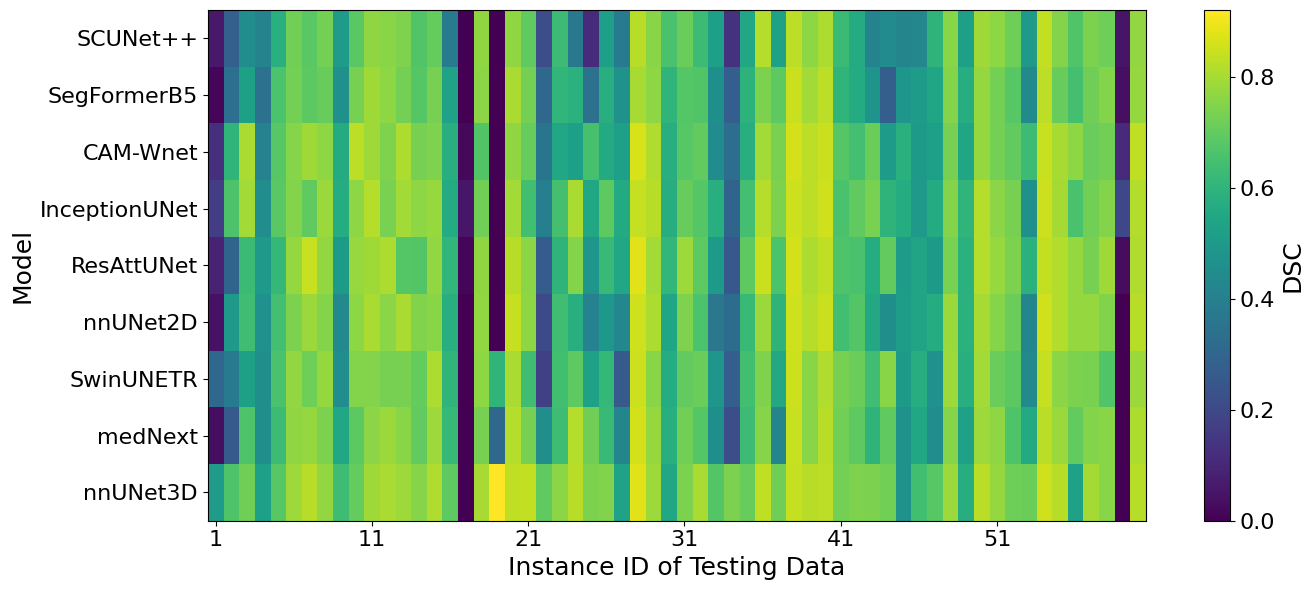}
    \caption{Vectors of DSC scores achieved by each model on different patients in the test set.}
    \label{fig:DSC_sequence}
\end{figure}

Beyond comparing the aggregated test performance of each individual model, we also assess whether the models' performance on individual patients is normally and randomly distributed. This allows us to identify and uncover some failure modes in the current pipeline. For each model, we arrange the DSC scores for each patient in a predefined order to construct a 60-dimensional vector. This 60-dimensional vector is the model’s performance spectrum (Figure \ref{fig:DSC_sequence}). 

\begin{figure}[h!]
    \centering
    \includegraphics[width=.75\linewidth]{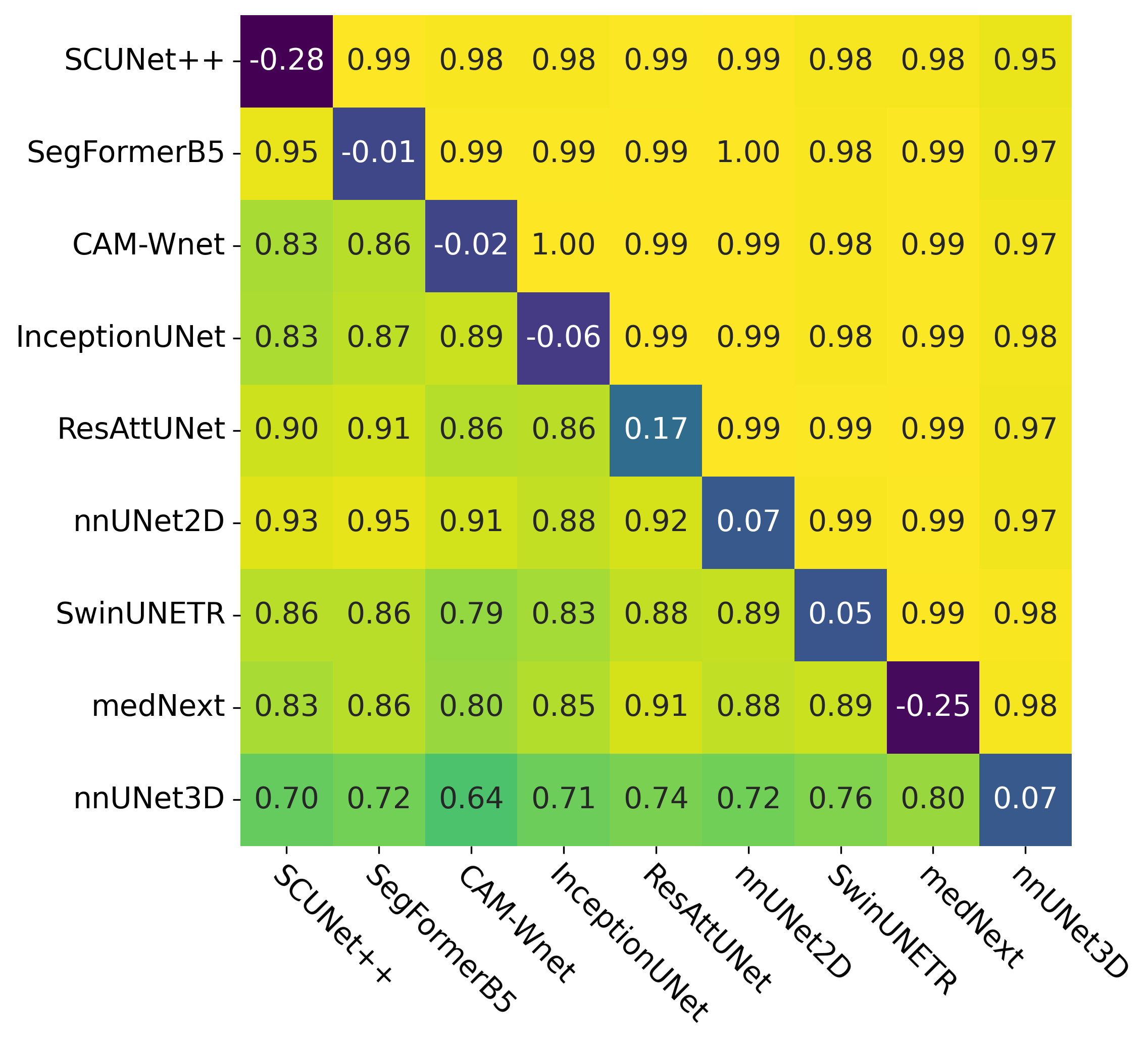}
    \caption{Upper Triangular: cosine similarity between DSC vectors; Lower Triangular: spearman R correlation between DSC vectors; Diagonal: spearman R correlation between two random positive vectors.}
    \label{fig:pred_sim_and_corr}
\end{figure}

Surprisingly, despite each model being randomly initialized, the distribution of their performance on the same test set no longer appears random after training. Specifically, when analyzing the performance spectra across different models, we observe high cosine similarity and Spearman rank correlation, indicating strong agreement among the performance spectra (Figure \ref{fig:pred_sim_and_corr}). It is also worth mentioning that, based on the spectra, all models exhibit false-negative predictions (DSC = 0) on a few patients, which drag down the aggregated mDSC. A closer examination of the false-negative cases may help unveil the roots, and we conduct this analysis in Section \ref{Sec: PE_detection_Rslt} in the context of PE detection.

\subsection{Embolus-level Detection Results}
\label{Sec: PE_detection_Rslt}
\label{sec: rslt_generalizability}
\subsubsection{Detection on In-House Dataset}
On our in-house test set, as reported in Table \ref{tab:generalization_comparison}, the highest-performing model achieved 181 true positives (TP), 49 false positives (FP), and 28 false negatives (FN) under a commonly used detection criterion (1-pixel overlap). 
As the detection threshold increased, the number of true positives decreased, while false negatives increased correspondingly; false positives exhibited a modest upward trend. 
Despite these changes, the decline in true positives was gradual, suggesting that the model’s predictions remain relatively stable across varying thresholds. Qualitative examples of TP, FP, and FN cases are shown in Figure \ref{fig:pe-visualization}.

\begin{figure}[h!]
    \centering
    \includegraphics[width=.75\linewidth]{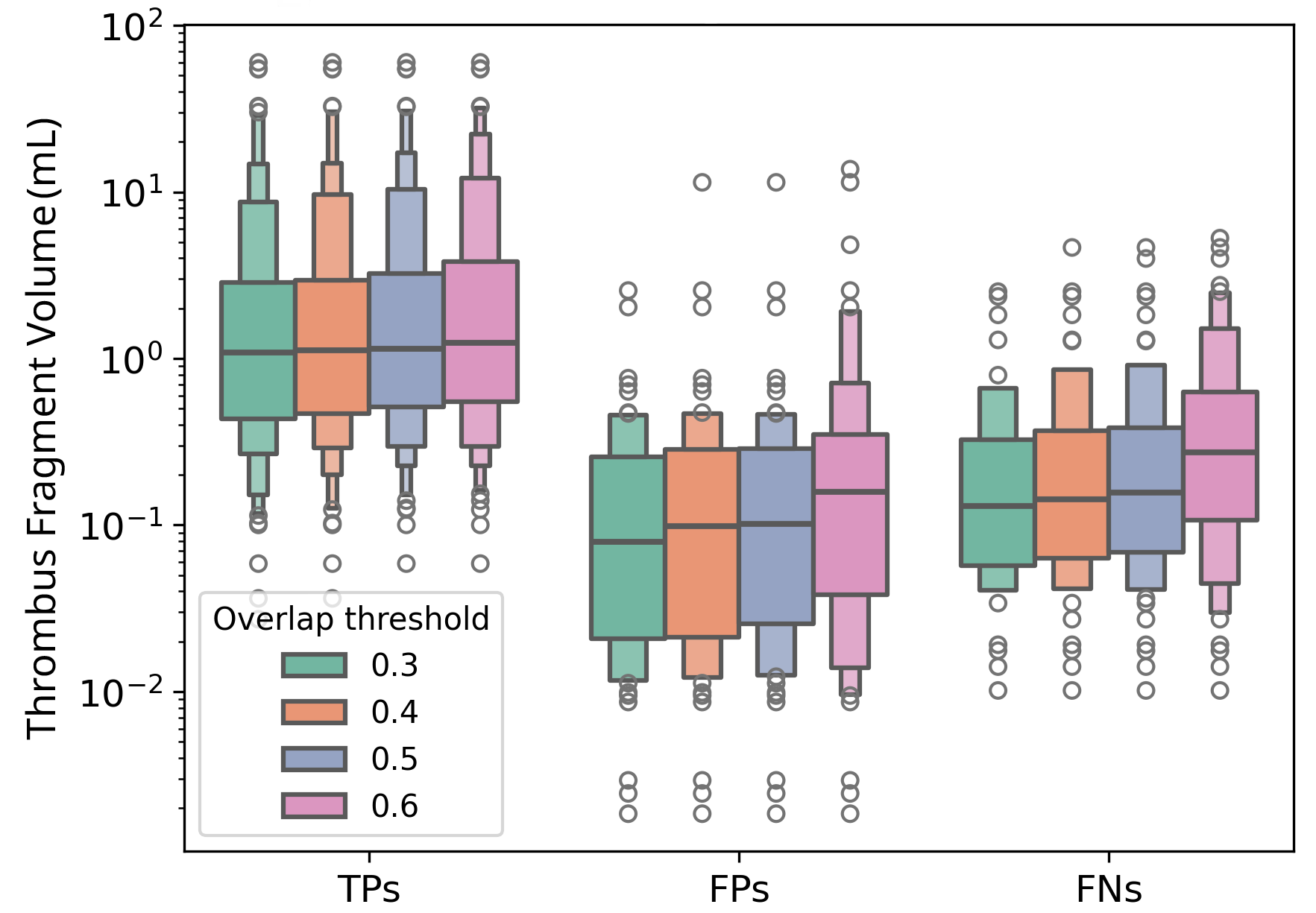}
    \caption{Boxen plots of fragment sizes by detection outcome on in-house dataset.}
    \label{fig:pe-size-distribution}
\end{figure}

\begin{table}[h!]
    \centering
    \begin{tabular}{|c|c|c|c|}
        \toprule
        \textbf{Detect Tr.} & \textbf{\# TP} & \textbf{\# FP} & \textbf{\# FN} \\
        \midrule
        1px & 181 & 49 & 28 \\
        0.1 & 179 & 49 & 30 \\
        0.2 & 173 & 50 & 36 \\
        0.3 & 170 & 52 & 39 \\
        0.4 & 161 & 56 & 48 \\
        0.5 & 154 & 59 & 55 \\
        0.6 & 138 & 68 & 71 \\
        \bottomrule
    \end{tabular}
    \caption{Thrombus-level detection results across different success criteria for outputs from the highest-performing model on our in-house test set (60 scans).}
    \label{tab:generalization_comparison}
\end{table}
It is worth noting that the distribution of prediction errors is skewed toward small distal PEs or artifacts. Figure \ref{fig:pe-size-distribution} illustrates the distribution of thrombus fragment volumes by detection outcome, showing that both false positives and false negatives predominantly occur among small distal fragments.

\subsubsection{Detection on Public Datasets}
\begin{figure*}
    \centering
    \includegraphics[width=0.325\linewidth]{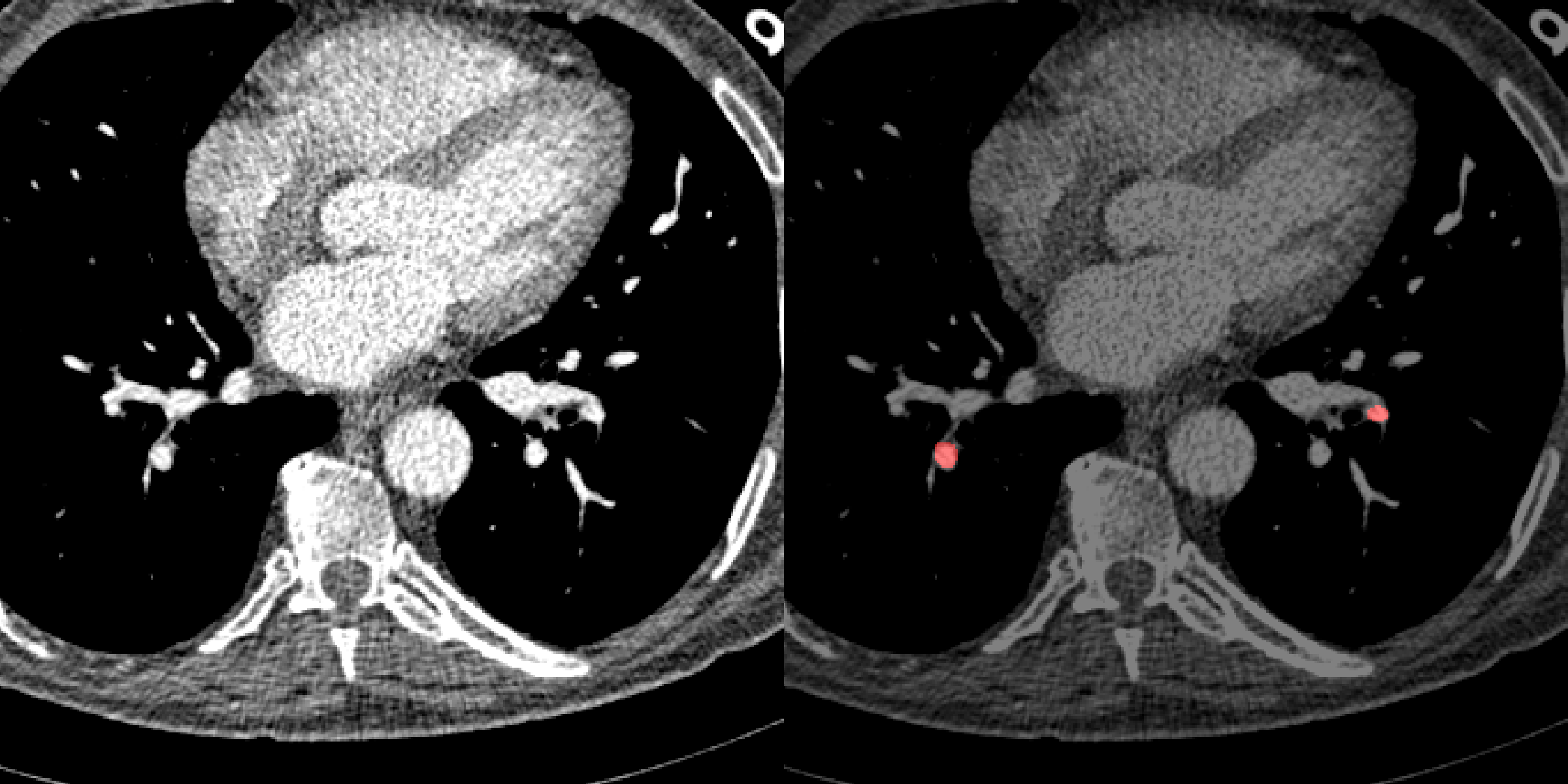}
    \includegraphics[width=0.325\linewidth]{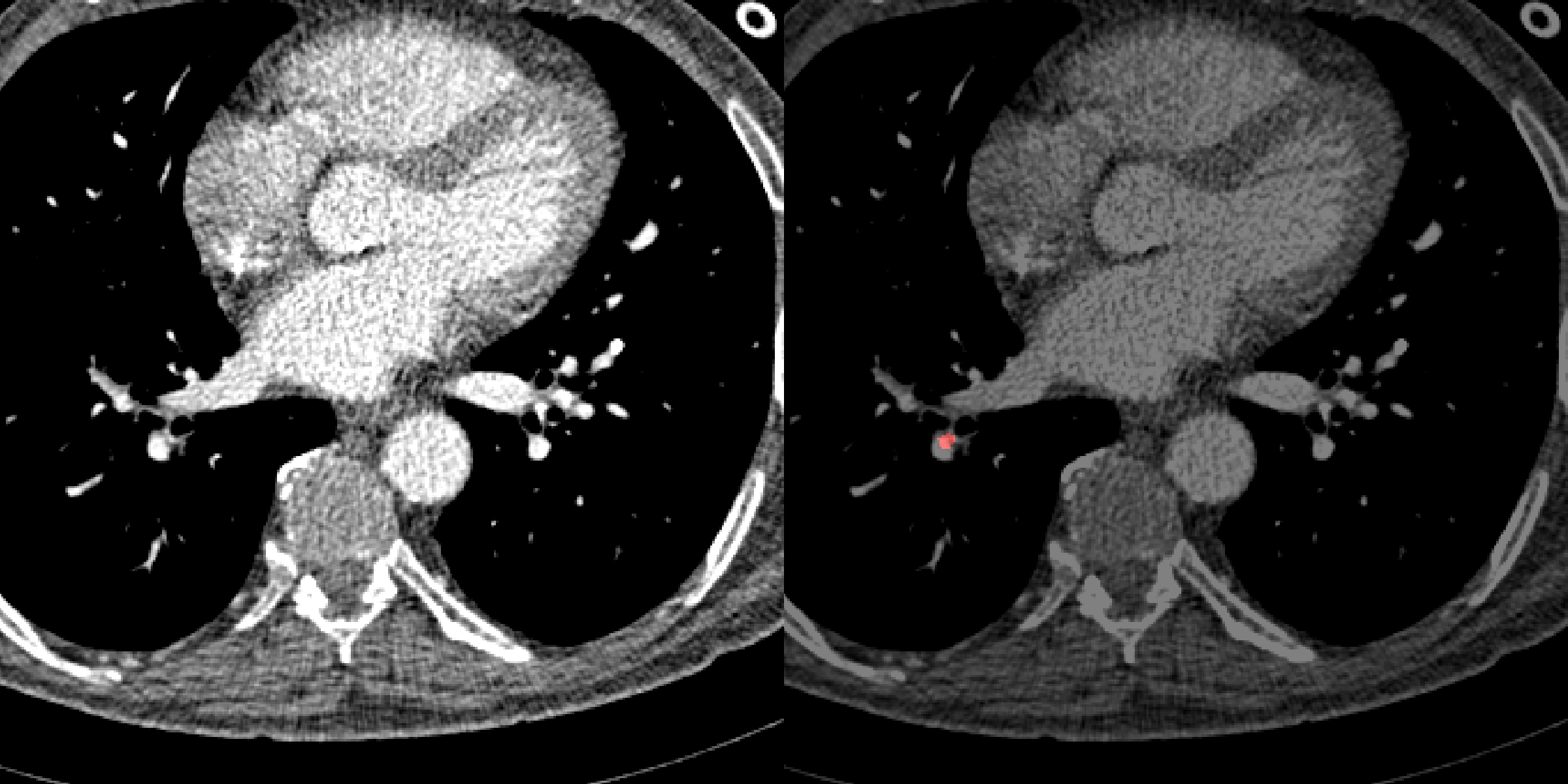}
    \includegraphics[width=0.325\linewidth]{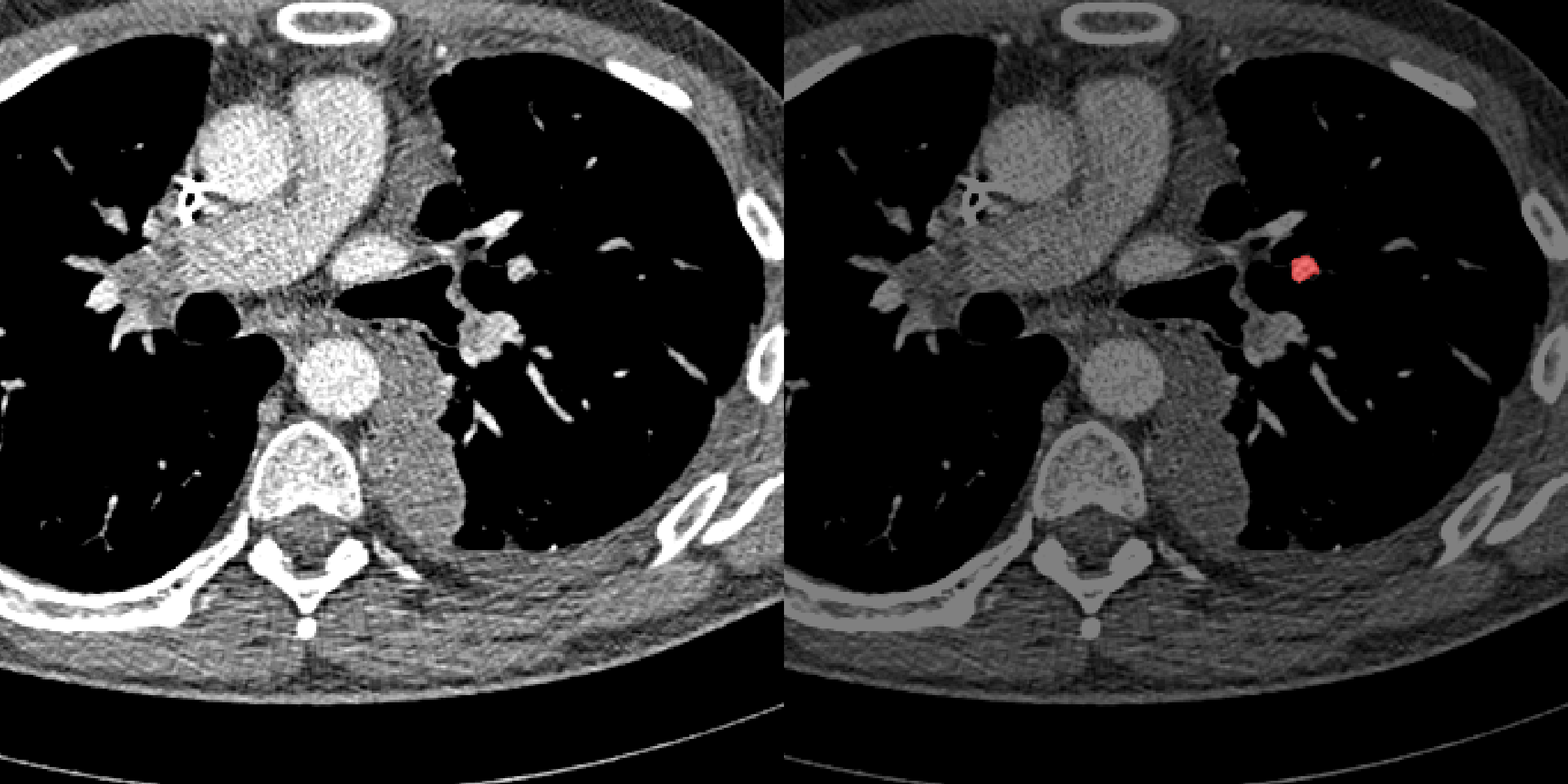}
    \caption{CAD-PE frequently provides annotations for the entire vessel or a large "dot" containing the PE, rather than delineating the embolus itself. In addition, some annotations correspond to low-contrast, PE-like regions that persist across multiple slices with minimal variation, raising concerns about their specificity. In certain cases, the annotations also appear to capture the interface between adjacent structures, such as the boundary between a vessel and an airway, rather than true intraluminal emboli.}
    
    \label{fig:CADPE_vis}
\end{figure*}

\begin{figure*}
    \centering
    \includegraphics[width=0.325\linewidth]{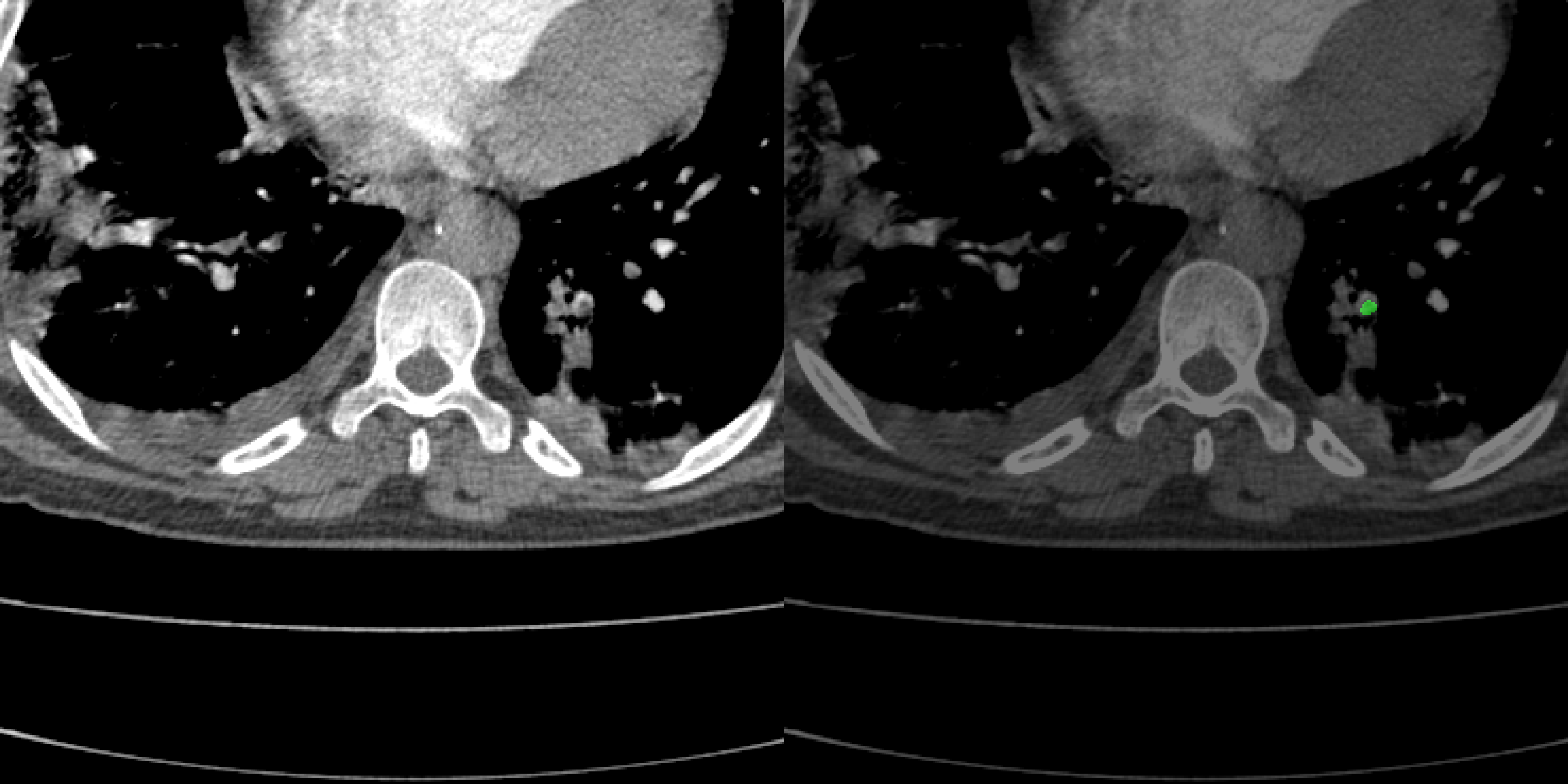}
    \includegraphics[width=0.325\linewidth]{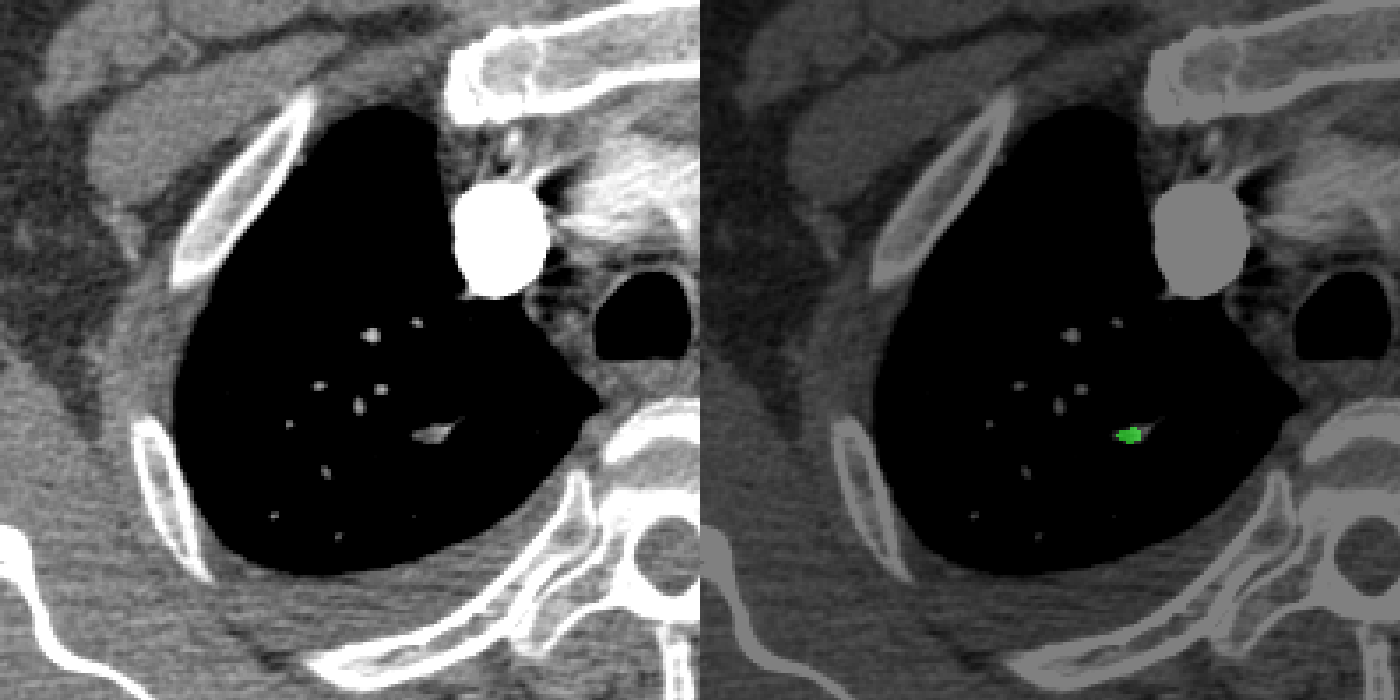}
    \includegraphics[width=0.325\linewidth]{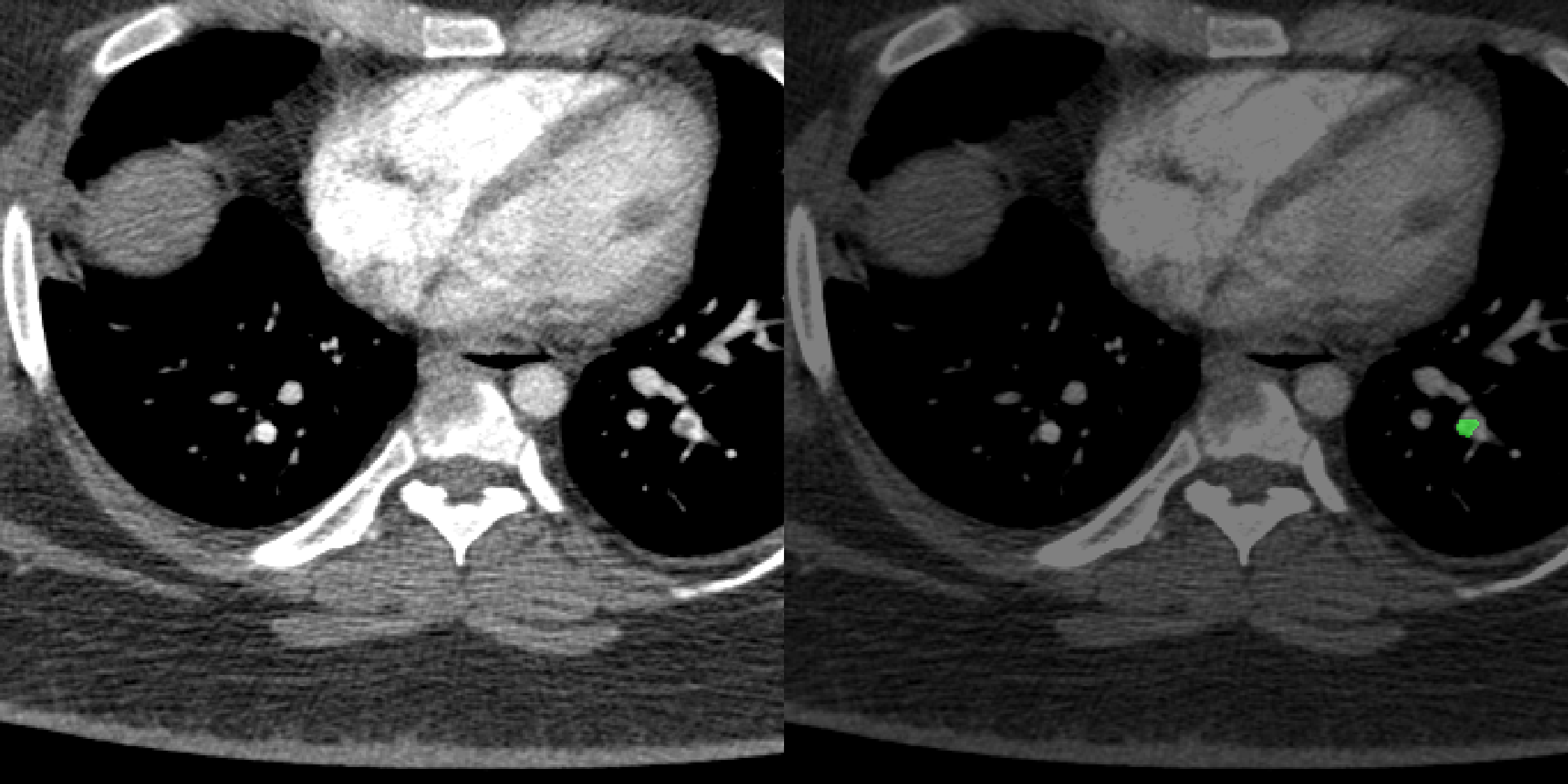}
    \caption{FUMPE lacks segmentation annotations for distal PEs which, however, are correctly identified by our highest-performing model. (Zoom-in for more details)}
    \label{fig:FUMPE_Vis}
\end{figure*}

\begin{figure*}
    \centering
    \includegraphics[width=0.325\linewidth]{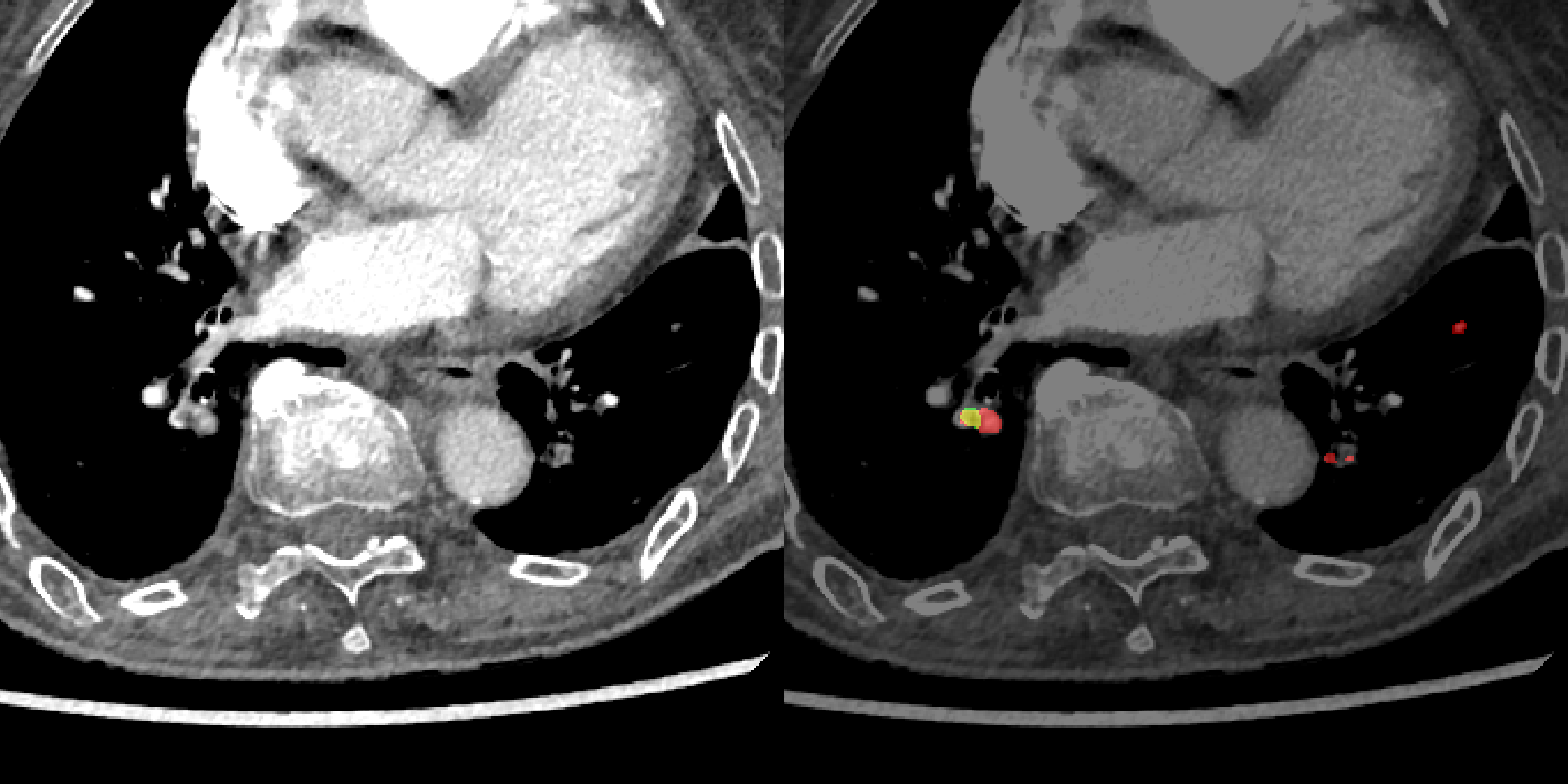}
    \includegraphics[width=0.325\linewidth]{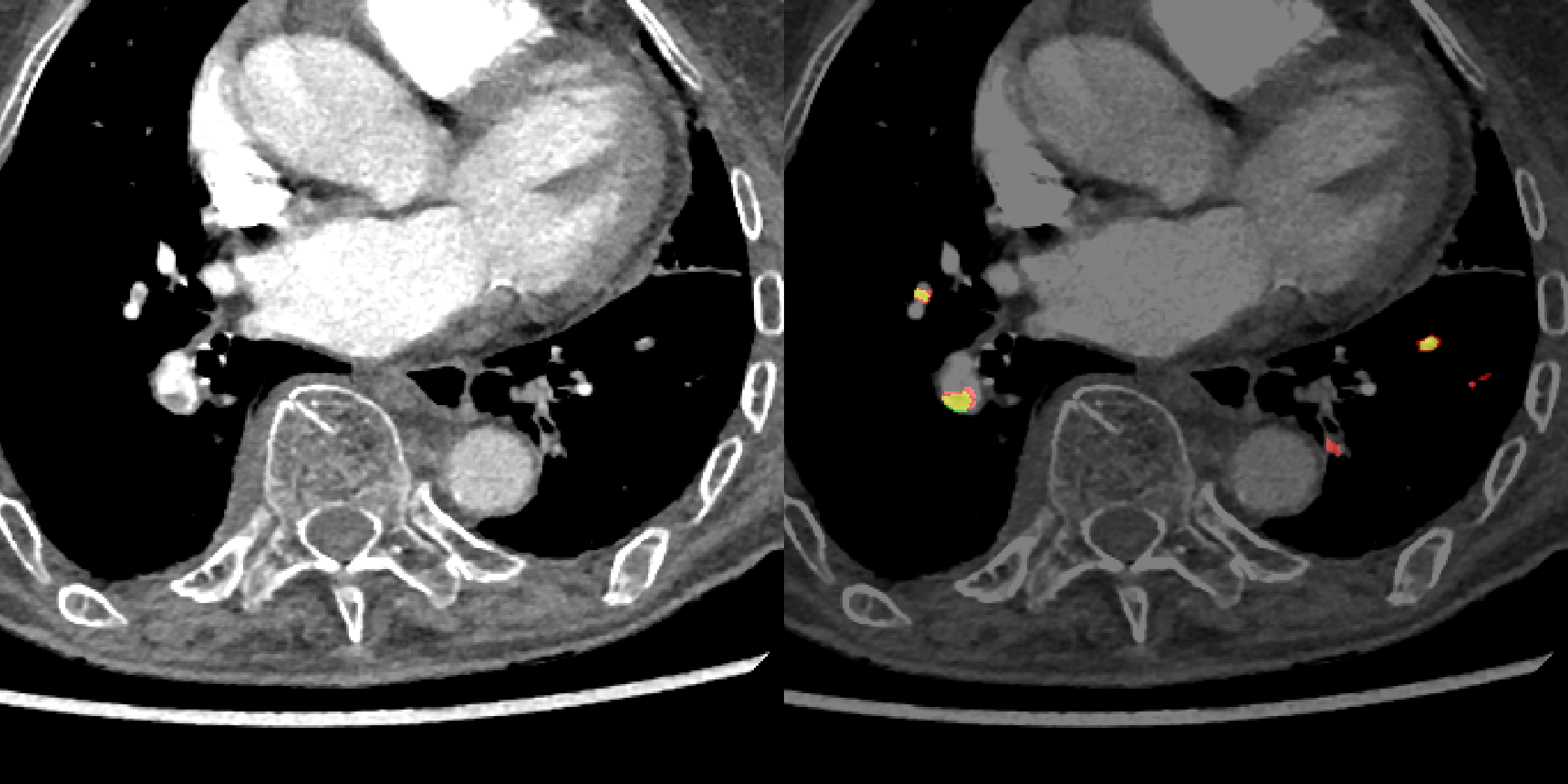}
    \includegraphics[width=0.325\linewidth]{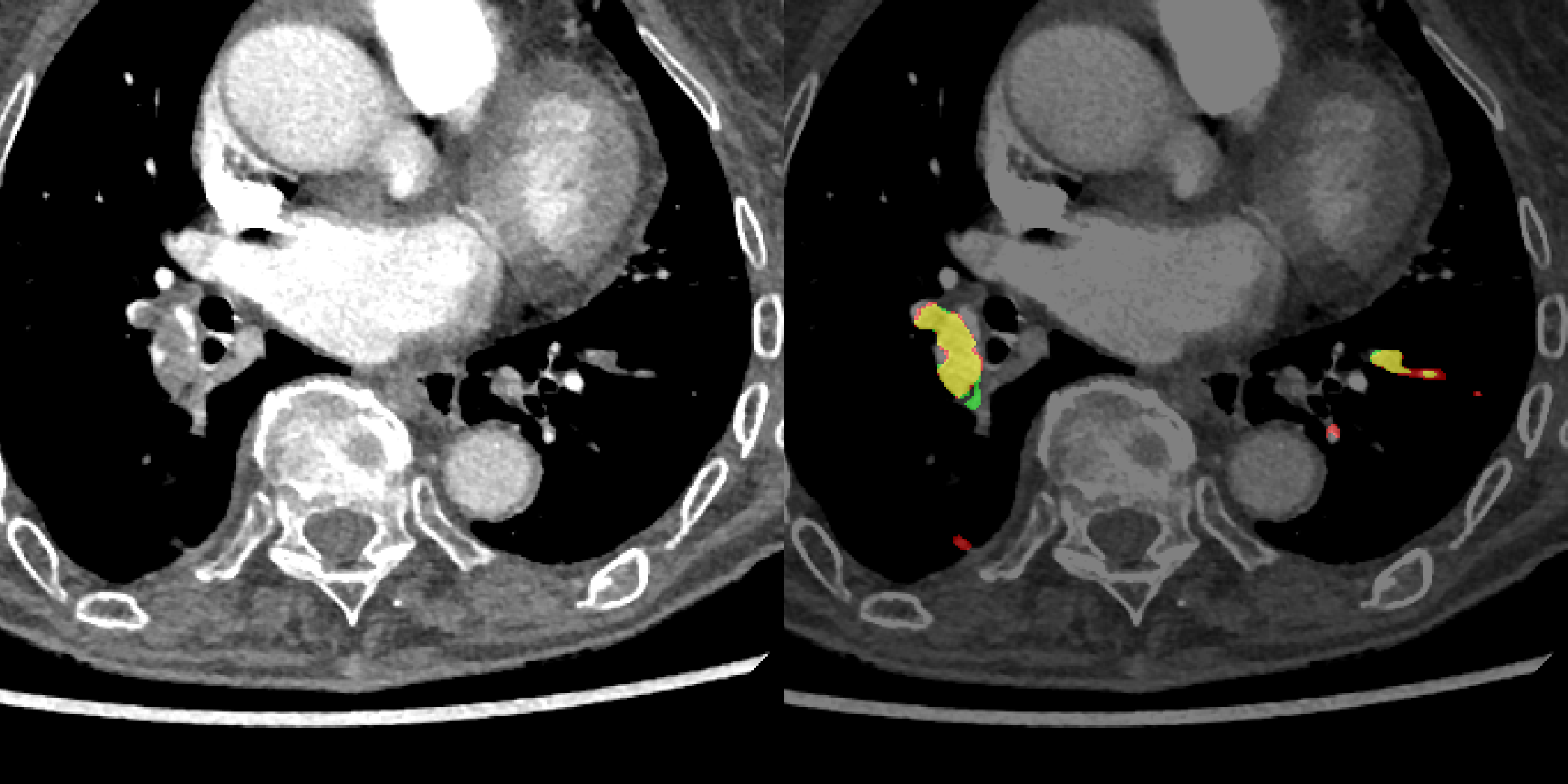}
    \caption{The READ annotations contain small, scattered artifacts that appear only on isolated slices.}
    
    \label{fig:READ_vis}
\end{figure*}

At first glance, the detection performance on public datasets appears less favorable than that observed on our in-house dataset. \textcolor{black}{However, a closer examination of the correspondence between the provided segmentation masks and the model’s predictions (see Fig. \ref{fig:CADPE_vis}, \ref{fig:FUMPE_Vis}, and \ref{fig:READ_vis}) reveals that differences in annotation quality and PE inclusion criteria play a substantial role. 
Consistent with the observations discussed} in Sections \ref{sec:intro} and \ref{sec:public_validation}, as well as in Fig. \ref{fig:thrombus_size_dist}, these results are largely explainable. Specifically, FUMPE excludes sub-segmental PEs from its annotations, \textcolor{black}{which can lead to correctly predicted distal emboli being counted as false positives, thereby inflating the FP rate.} In contrast, CAD-PE includes annotations that appear to correspond to partial-volume artifacts rather than true emboli, which may contribute to an elevated FN rate. \textcolor{black}{Importantly, when comparing the FN rate for FUMPE and the FP rate for CAD-PE (Tables \ref{tab:generalization_comparison} and \ref{tab:combined_baseline_comparison}), these metrics are comparable to the corresponding values observed on our in-house dataset.} \textcolor{black}{For results at higher thresholds, we report them in Table \ref{tab:combined_baseline_comparison_higherTr} in the Appendix, as the visualizations suggest that performance differences at this stage are dominated by disparities in annotation quality rather than true differences in model capability.}

\begin{table}[h!]
    \centering
    \begin{tabular}{|c|c|c|c|c|}
        \toprule
        \textbf{Dataset} & \textbf{Detect Tr.} & \textbf{\# TP} & \textbf{\# FP} & \textbf{\# FN} \\
        \midrule
        FUMPE & 1px & 77  & 60 & 16  \\
        (35 scans) & 0.1 & 71  & 61 & 22  \\
                   & 0.2 & 70  & 61 & 23  \\
                   & 0.3 & 68  & 61 & 25  \\

        \midrule
        CAD-PE & 1px & 215 & 15 & 168 \\
        (91 scans) & 0.1 & 186 & 17 & 197 \\
                   & 0.2 & 149 & 17 & 234 \\
                   & 0.3 & 113 & 17 & 270 \\

        \midrule
        READ  & 1px & 115 & 27 & 78  \\
        (40 scans) & 0.1 & 104 & 27 & 89  \\
                   & 0.2 & 98  & 28 & 95  \\
                   & 0.3 & 84  & 28 & 109 \\
                   
        \bottomrule
    \end{tabular}
    \caption{Thrombus-level detection results across different success criteria for outputs from the highest-performing model on the \textbf{public datasets} (FUMPE, READ, and CADPE).}
    \label{tab:combined_baseline_comparison}
\end{table}

\begin{figure}[h!]
    \centering
    \includegraphics[width=.75\linewidth]{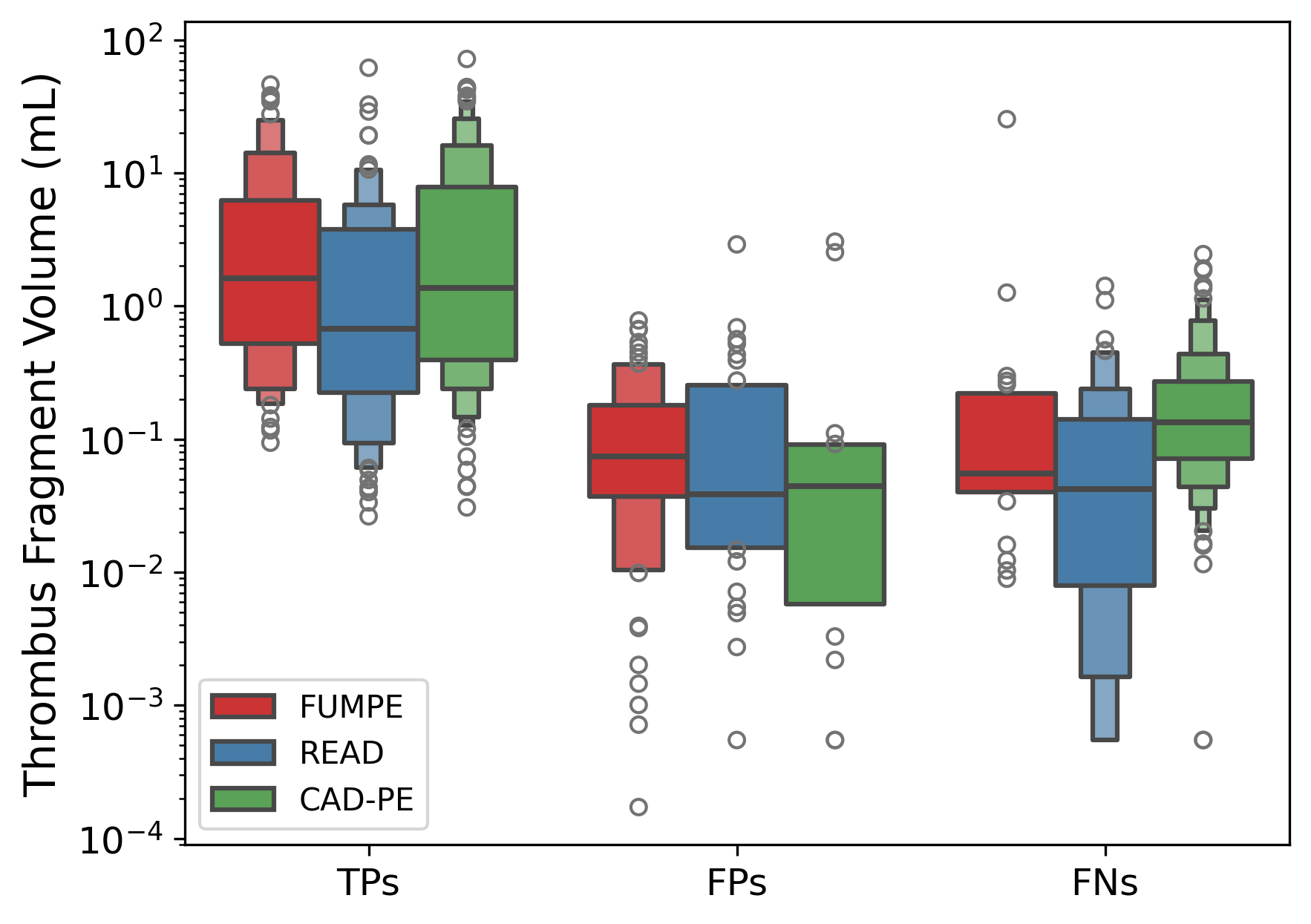}
    \caption{Boxen plots of fragment sizes by detection outcome on public datasets.}
    \label{fig:public-size-distribution}
\end{figure}

READ consists of near-isomorphic, high-resolution CTPA scans with mostly precise annotations. On this dataset, our model achieves a false positive (FP) rate comparable to that observed on our in-house dataset. The higher false negative (FN) rate, although seemingly counterintuitive, may be attributable to the high-resolution annotations provided by READ. In Fig. \ref{fig:public-size-distribution}, we present boxen plots of thrombus fragment size distributions stratified by detection outcome for each public dataset. Notably, the FN cases in READ exhibit a long-tailed distribution toward very small fragment volumes. Specifically, more than 33.3\% of the missed embolic fragments in READ have volumes smaller than 0.01 mL, which is typically below the spatial resolution required for reliable visualization or annotation in routine CTPA. \textcolor{black}{It is also possible that certain anatomical details are lost during the resampling process.} This observation helps explain why our model achieves similar average TP and FP counts but a higher average FN count on READ. \textcolor{black}{Qualitative inspection further shows that many of the missed detections correspond to small, scattered annotations}.

\section{Discussion}
Our study reveals several patterns that may inform future research on pulmonary embolism (PE) segmentation. Across all model architectures, segmentation performance was highly correlated at the patient level, \textcolor{black}{suggesting that model-specific architectural differences may play a secondary role compared to data-related factors in determining performance.} Further failure analysis showed that the median size of both false-positive and false-negative fragments is more than an order of magnitude smaller than that of the true-positive counterparts. This finding is consistent with the non-technical explanation proposed by Zhou et al. \cite{zhou2019variabilities}, which suggests that even experienced radiologists struggle to precisely identify sub-segmental PEs and are prone to false positives during annotation. \textcolor{black}{Together, these results indicate that the primary challenge of PE segmentation lies in reliably detecting small and ambiguous emboli.}

\textcolor{black}{Given these observations, we hypothesize that increasing the number of clinically confirmed difficult cases by several-fold would be a key driver for improving model performance on this problem.} \textcolor{black}{In parallel, methodological improvements should focus on enhancing spatial context understanding.} Multi-view or volumetric perception mechanisms should be actively considered when developing or deploying PE segmentation models, as small PEs can be invisible in certain plane directions, \textcolor{black}{while suspected PEs can be more reliably differentiated from partial-volume artifacts when richer spatial context is available.} \textcolor{black}{These findings collectively suggest that both data-centric and architecture-centric strategies are necessary to address the limitations observed in current models.}

To our surprise, classification is not demonstrated to be an effective pretext task for pre-training when \textcolor{black}{the final objective is PE segmentation.} \textcolor{black}{This indicates a notable discrepancy between the features learned for classification and those required for precise segmentation.} This is reasonable because, for classification, the model may rely more on the anatomical context surrounding the PE (as suggested by Grad-CAM visualizations in prior PE slice-level classification studies), whereas for small structures such as PEs, the feature representation may fail to preserve sufficiently localized information for accurate boundary delineation, which is critical for segmentation, leading to minor performance degradation. \textcolor{black}{Another possible explanation is that our curated training set containing 430 CTPAs already captures substantial anatomical variability that can be inferred from the RSNA-PE dataset \cite{RSNA-PE}. In this case, further improvements may depend less on general representation learning and more on explicitly targeting difficult cases. However, such stratification strategies have not yet been implemented in any existing studies.}

Regarding the three public datasets for PE segmentation, \textcolor{black}{qualitative visualization of the results in Figures \ref{fig:CADPE_vis}, \ref{fig:FUMPE_Vis}, and \ref{fig:READ_vis} (FUMPE, CAD-PE, and READ) reveals varying degrees of annotation error, imprecision, and mismatch between labels and underlying images.} \textcolor{black}{These inconsistencies may introduce additional noise during training and evaluation, potentially limiting model performance and generalizability.} A revision, stratification, or annotation augmentation of existing public datasets could represent a meaningful contribution that would benefit both PE segmentation research and the broader medical AI community. \textcolor{black}{In contrast, our in-house dataset demonstrates improved annotation quality and consistency, which translates into superior model performance. We anticipate that the models and pretrained weights derived from this dataset can serve as a valuable resource to accelerate future efforts in PE segmentation.}

\section{Conclusions}
Our study conducts a comprehensive audit of the existing literature on pulmonary embolism segmentation \textcolor{black}{algorithms, public datasets, and evaluation pipelines. We further curated a PE segmentation dataset comprising} 490 unique patients and release an open-weight model trained on this dataset to support the research community. \textcolor{black}{The primary limitation of our work is that the dataset consists predominantly of routine clinical CTPA scans, and the model is designed for such imaging characteristics; consequently, it may be less suitable for near-isomorphic, high-resolution CTPA (which are less common in clinical practice), where resampling is required and may lead to the loss of fine anatomical detail. Additionally, as the dataset is derived from human interpretation, it may contain inaccuracies associated with radiologist variability and error. Future work will investigate whether radiomics features and biomarkers derived from segmentation outputs align with clinical reality and improve model generalizability. Furthermore, the development of public datasets with precise, embolus-level annotations of distal PEs will require substantial contributions and rigorous validation by clinical experts.}     

\section*{Acknowledgments}
Research reported in this publication was supported by the National Heart, Lung, and Blood Institute of the National Institutes of Health under Award Number R44HL152825. The content is solely the responsibility of the authors and does not necessarily represent the official views of the National Institutes of Health.

\bibliographystyle{unsrt}
\bibliography{main}

\appendix
\color{black} 
\section{Manufacturer and Imaging Parameter of the In-House dataset}
\label{Image_param}
Our in-house dataset was acquired using a diverse set of CT scanners, with detailed manufacturer statistics provided in Table~\ref{tab:ct_scanner_stats}. 
The kVp and mAs values were extracted from the \texttt{Exposure} field of the original DICOM files for all manufacturers except GE. For GE scanners, the \texttt{Exposure} field did not contain informative mAS values. In those cases, mAS was derived from other DICOM attributes according to:
$$mAS = \dfrac{XRayTubeCurrent\times TableFeedPerRotation}{TableSpeed\times SpiralPitchFactor} $$

\begin{table}[htbp]
\centering
\begin{tabular}{lrrr}
\hline
\textbf{Manufacturer} & \textbf{N} & \textbf{kVp} & \textbf{mAs} \\
\hline
Canon & 1   & 120.0 (N/A) & 225.0 (N/A)  \\
NMS & 3   & 113.3 (11.6) & 248.0 (127.5) \\
Hitachi & 4   & 115.0 (10.0) & 316.0 (148.2) \\
Philips & 58  & 115.2 (9.4)  & 281.3 (137.3) \\
Toshiba & 64  & 115.3 (10.9) & 189.1 (76.5) \\
Siemens & 114 & 115.6 (11.1) & 193.4 (84.0) \\
GE & 246 & 118.1 (9.2)  & 235.1 (96.4) \\
\hline
Total & 490 & 116.7 (10.0) & 225.6 (101.7)\\
\hline
\end{tabular}
\caption{CT Scanner Manufacturer Distribution and Acquisition Parameters}
\label{tab:ct_scanner_stats}
\end{table}

\section{Statistical results for model performance comparison}
\begin{figure}[h!]
    \centering
    \includegraphics[width=.75\linewidth]{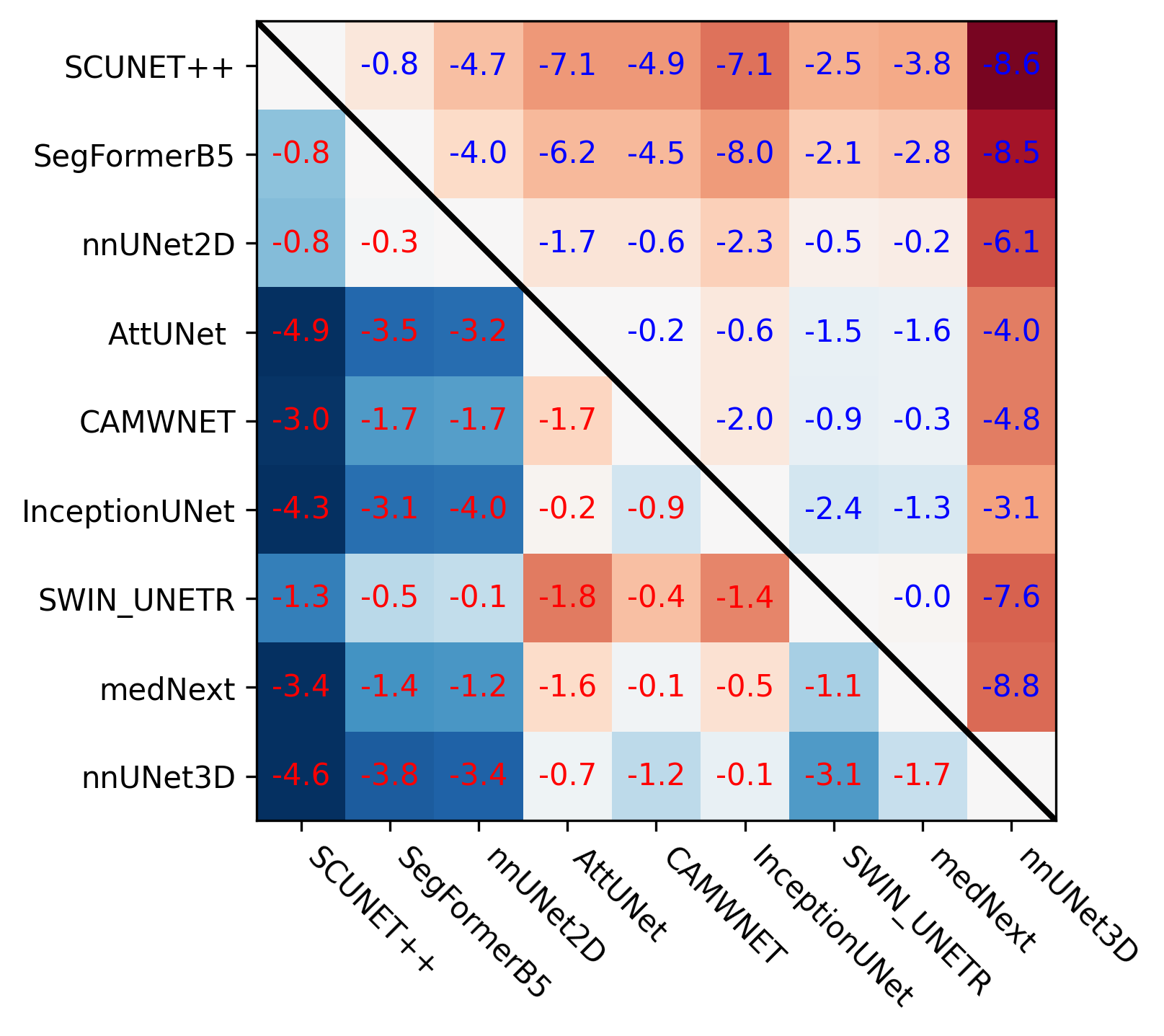}
    \caption{The p-value for model performance differences after the logarithm transformation with base 10. Upper triangular shows DSC and lower triangular shows ASSD (blue/red hue means model in the row is higher/lower-performing than model in the column).Darker color means larger difference in average performance.}
    \label{fig:p_value}
\end{figure}

We conduct statistical analysis to evaluate model performance and determine whether statistically significant differences exist between different model architectures. Since the segmentation results did not satisfy the assumption of normality, we employ the Wilcoxon signed-rank test rather than the pair-wise t-test for comparisons. The p-value after the logarithmic transformation with base 10 is plotted as text in Figure \ref{fig:p_value}. A statistically significant difference in model performance is confirmed by the low logarithmic value of the p-value. 

\section{PE detection on higher threshold}
\begin{table}[h!]
    \centering
    \begin{tabular}{|c|c|c|c|c|}
        \toprule
        \textbf{Dataset} & \textbf{Detect Tr.} & \textbf{\# TP} & \textbf{\# FP} & \textbf{\# FN} \\
        \midrule
        FUMPE & 0.4 & 56  & 61 & 37  \\
        (35 scans) & 0.5 & 43  & 63 & 50  \\
                   & 0.6 & 33  & 65 & 60  \\
                
        \midrule
        CAD-PE & 0.4 & 74 & 18 & 309 \\
        (91 scans) & 0.5 & 32 & 21 & 351 \\
                   & 0.6 & 12 & 22 & 371 \\
        
        \midrule
        READ  & 0.4 & 77 & 28 & 116  \\
        (40 scans) & 0.5 & 58 & 31 & 135  \\
                   & 0.6 & 38  & 37 & 155  \\
                   
        \bottomrule
    \end{tabular}
    \caption{Thrombus-level detection results across different success criteria for outputs from the highest-performing model on the \textbf{public datasets} (FUMPE, READ, and CADPE).}
    \label{tab:combined_baseline_comparison_higherTr}
\end{table}

\end{document}